\documentclass[runningheads]{llncs}
\usepackage{graphicx}
\usepackage{comment}
\usepackage{amsmath,amssymb} 
\usepackage{color}

\usepackage{cite}

\begin{document}
\pagestyle{headings}
\mainmatter
\def\ECCVSubNumber{4809}  

\title{Filter Style Transfer between Photos} 

\titlerunning{Filter Style Transfer between Photos}
%
\author{Jonghwa Yim \and
Jisung Yoo\thanks{Corresponding Author} \and
Won-joon Do \and
Beomsu Kim \and
Jihwan Choe}
\authorrunning{Jonghwa Yim et al.}
%
\institute{Visual Solution Lab., Samsung Electronics, South Korea\\
\email{\{jonghwa.yim, jisung.yoo, wonjoon.do, bs8207.kim, jihwan.choe\}@samsung.com}}
\maketitle

\begin{figure}
\centering
\includegraphics[width=\linewidth]{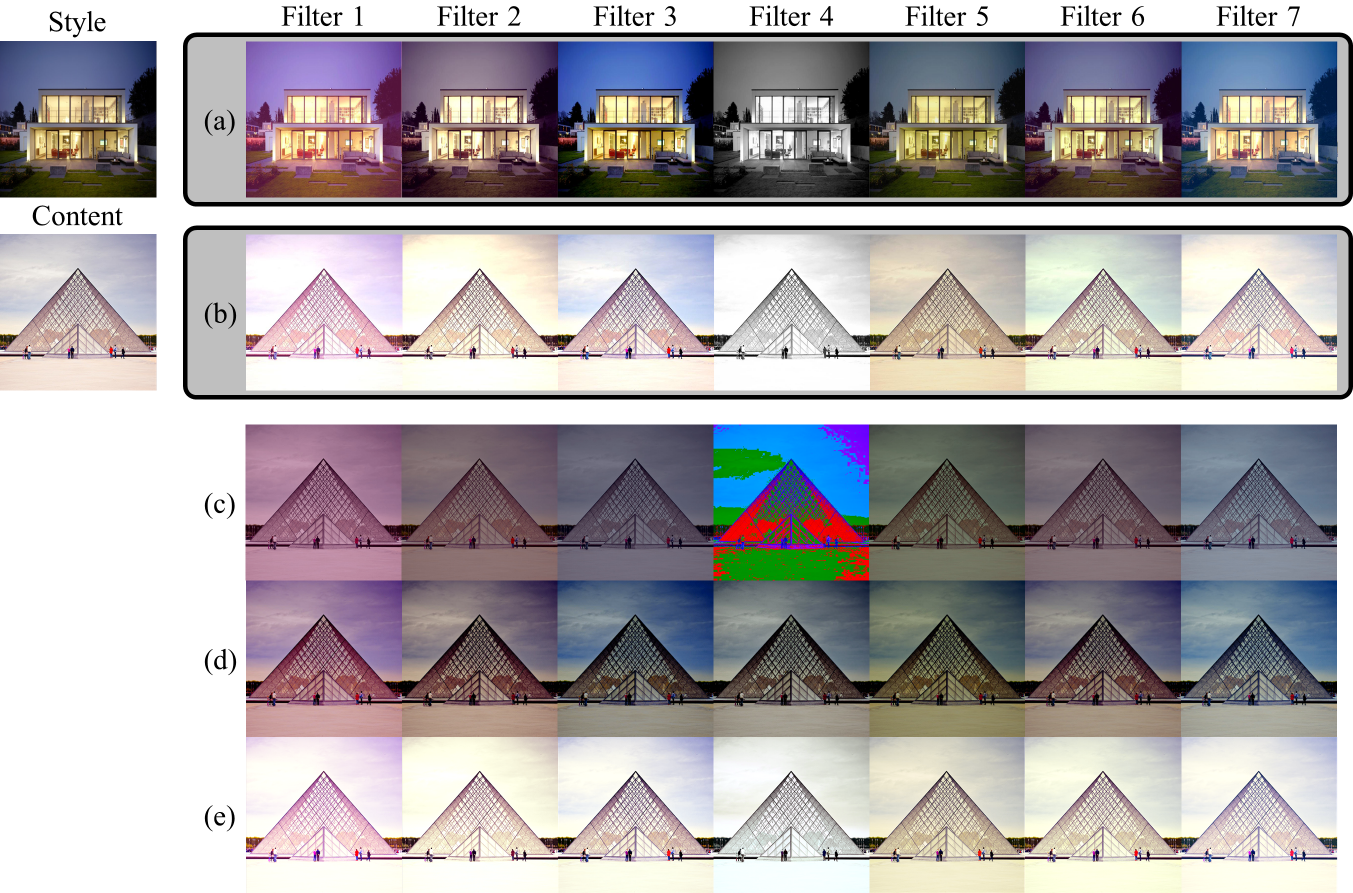}
\caption{{\bf Filter style transfer results.} Given reference images with arbitrary filters applied (a), the filter styles can be transferred to a new image with our model. While (b) is the ground-truth, (c)-(e) show the results of color transfer~\cite{reinhard2001color}, photorealistic style transfer: WCT2~\cite{yoo2019photorealistic}, and ours, respectively.}
\label{fig:fig1}
\end{figure}

\begin{abstract}
Over the past few years, image-to-image style transfer has risen to the frontiers of neural image processing. While conventional methods were successful in various tasks such as color and texture transfer between images, none could effectively work with the custom filter effects that are applied by users through various platforms like Instagram. In this paper, we introduce a new concept of style transfer, Filter Style Transfer (FST). Unlike conventional style transfer, new technique FST can extract and transfer custom filter style from a filtered style image to a content image. FST first infers the original image from a filtered reference via image-to-image translation. Then it estimates filter parameters from the difference between them. To resolve the ill-posed nature of reconstructing the original image from the reference, we represent each pixel color of an image to class mean and deviation. Besides, to handle the intra-class color variation, we propose an uncertainty based weighted least square method for restoring an original image. To the best of our knowledge, FST is the first style transfer method that can transfer custom filter effects between FHD image under 2ms on a mobile device without any textual context loss.
\end{abstract}

\section{Introduction}

Stylizing an image with characteristics of other stylized images has long been a difficult problem in Computer Vision. Beyond simple editings, people's desire to grand artistic feelings to their pictures has increased. For this reason, a tool that can stylize their photos in a unique way is highly desired.

There have been several studies addressing technical solutions for image-to-image style or content transfer. Reinhard et al.\cite{reinhard2001color} is one of the pioneering attempts where mean and variance of RGB color distribution from a source image were used to apply the color scheme to a target image, but with limited success obtaining enough similarity between two images. Others~\cite{welsh2002transferring, xiao2006color, tai2005local, pitie2007automated} tried to improve results by using various mathematical approaches to treat color distribution but failed to consider the semantics of pictures during the process. Moreover, their methods transferred objects’ inherent colors as well, limiting their methods in the assumption that scene components of the two images must be similar.

More recently, taking advantage of the advent of Deep Neural Network, more sophisticated applications of image-to-image style transfer became possible. Style transfer~\cite{gatys2016image} encoded not only color but also shapes and textures. After that, many following works branched out to further improve the accuracy and efficiency of the style transfer. Some researches~\cite{isola2017image, liu2017unsupervised, zhu2017toward} were related to domain transfer, which transfers styles between different image domains such as semantic-labels to street-scene, aerial to map, and sketch to photo. However, since most domain transfer approaches aimed to move input images' distributions close to the target domain, the output of them does not explicitly reflect the style of a single image. Some other researches~\cite{luan2017deep, li2018closed, sheng2018avatar, yoo2019photorealistic} introduced methods to transfer photorealistic styles from a single style image, but they required a large dataset for training leading to a high computational cost. Even with the high processing time, they often displayed undesirable transfers of colors and textures due to fundamentally implicit actions of deep neural networks. In this case, it is difficult to identify causes and solutions, which is a significant hurdle when commercializing the approaches.

Meanwhile, some researches suggested automated photograph editing to enhance the overall quality \cite{kang2010personalization, bychkovsky2011learning, kaufman2012content, yan2014learning, yan2016automatic, chandakkar2017joint, gharbi2017deep, ignatov2017dslr, chen2018deep, omiya2018learning, bianco2019learning, bianco2019content} or control exposure \cite{yuan2012automatic, yu2018deepexposure, yang2018personalized, hu2018exposure}. All of these researches showed considerable progress on automated image editing, but they also required large datasets to train an image enhancement model. More importantly, they only followed a predefined editing rule like High Dynamic Range (HDR). Some focused on the extraction of photo-editing-parameters directly \cite{gharbi2017deep, omiya2018learning, bianco2019learning, bianco2019content}, while \cite{chen2018deep} focused on learning image enhancement using GAN~\cite{goodfellow2014generative} to generate HDR output. A method in \cite{bianco2019learning} suggested parameter extraction from a neural network, but it was not a single-stage and showed limited performance. Also, efforts to model polynomial functions in the previous studies~\cite{bianco2019learning, bianco2019content} may suffer from high-order variables' fitting issues as well as they still limited their methods to predefined editing rules. There have been a few efforts to adopt reinforcement learning \cite{yu2018deepexposure, hu2018exposure} to train enhancement policies. Despite all the efforts, all the aforementioned methods were not able to extract filter parameters from an already stylized image and require the original version of the stylized image to enhance the target image. Such limitations prevented previous studies from fully satisfying commercial needs.

With the increased accessibility of mobile phones and the internet, these days, people spend even more time on social media. As a result, many photo-editing applications have been developed and are widely used with various stylizing filters to give special effects on photos taken. To the best of our knowledge, however, there has not been an attempt to extract custom filter effects from a stylized photo. In this study, a mathematical formulation of custom filter extraction and its application to new photos are presented. FST is quite efficient without requiring expensive computing time, even in a low-end mobile device; so the application can be easily adopted and used in our fast-moving social networking environment.

\section{Method Overview}

\begin{figure}
\centering
\includegraphics[width=\linewidth]{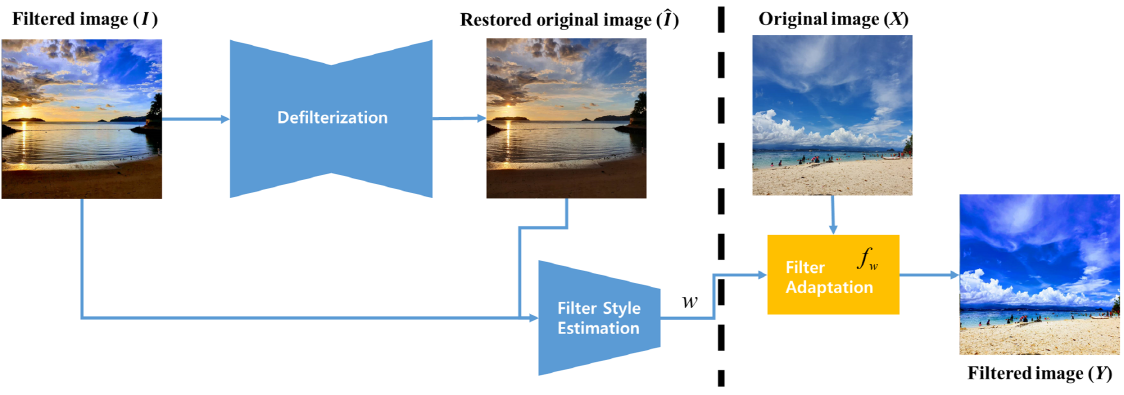}
\caption{{\bf System overview to extract the photo editing parameter and apply it to the new input images.} $I$ is a filtered or edited image that we want to extract filter style and $w$ represents the parameters of the filter style of $I$. The system is initially black-box and must be designed appropriately to infer standalone parameters.}
\label{fig:fig2}
\end{figure}

Fig.~\ref{fig:fig2} shows an overview of the proposed method in this research. It comprises extracting custom photo filters from a single reference image ($I$) and applying them to a new one ($X$). Our method restores the original image ($\hat{I}$) from $I$, which is called \textit{defilterization} in this paper. Then, using two images, a filter parameter $w$ is obtained, which is called \textit{filter style estimation}. Lastly, using $w$, a designed filter function $f_w$ can be used to filter the user's original image ($X$) to newly-stylized image ($Y$).

\section{Defilterization}

In Fig.~\ref{fig:fig2}, the stylized reference input $I$ is a projected image from the original image $\acute{I}$ using the filter-applying function $f_w$, leading to
\begin{align}
  I=f_w (\acute{I})\text{.}
\end{align}
To determine $f_w$, the relationship between the pair of the original image $\acute{I}$ and stylized image $I$ needs to be investigated. From (1), we know that
\begin{align}
  \acute{I}=f_w^{-1}(I)
\end{align}
Assume that there is a collection $S$ of $M$ stylized images, $S=\{I_1,I_2,…,I_M\}$. Each image $I$ consists of $K$ object segments, like the sky, cow, grass, etc., and each object segment can be represented its vectorized form $\boldsymbol{o}$, such that 
\begin{align}
  I = \{ \boldsymbol{o}_1,\boldsymbol{o}_2,\boldsymbol{o}_3,…,\boldsymbol{o}_K \}
\end{align}
where $o_k$ is a vector of colors of the flattened pixels in the $k$-th object segment. Then the original image of I can also be represented as a set of the original object segments, such that 
\begin{align}
  \acute{I} = \{ \acute{\boldsymbol{o}}_1, \acute{\boldsymbol{o}}_2, \acute{\boldsymbol{o}}_3,…,\acute{\boldsymbol{o}}_K \}
\end{align}
where $\acute{\boldsymbol{o}}_k$ is the original colors before stylized. If an implicit object, $\acute{\boldsymbol{o}}_k$, has some class label, $cls$, and its mean color can be obtained by averaging all pixel colors in $cls$ throughout the dataset $S$, then, $\acute{\boldsymbol{o}}_k$ can be expressed with the mean color value of the class, $\tilde{\boldsymbol{o}}_{k,cls}$, and pixel-wise color deviations, $\Delta_{k,cls}$. Note that we start to explain from the implicit object level to introduce the class label and its mean color. Also, there is a numerical error term $\epsilon_{k}$ due to imperfect restoration of original object colors. Therefore $\acute{\boldsymbol{o}}_k$ can be expressed as
\begin{align}
  \acute{\boldsymbol{o}}_k = \tilde{\boldsymbol{o}}_{k,cls} + \Delta_{k,cls} = f_w^{-1}(\boldsymbol{o}_k) + \boldsymbol\epsilon_k \text{.}
\label{eq:notations_obj_delta_eps}
\end{align}
The distance between the restored image, $f_w^{-1}(I)$, and the true original, $\acute{I}$, can be described as the sum of the squared distance between restored and true objects. 
\begin{align}
  Distance(f_w^{-1}(I), \acute{I}) & = \sum_{k=1}^{K}\|f_w^{-1}(\boldsymbol{o}_k) - \acute{\boldsymbol{o}}_k \|^2
\label{eq:distance}
\end{align}
Then our problem to find the original image becomes a minimization problem. At a pixel level, the objective function of the minimization process becomes
\begin{align}
  \operatorname*{arg\,min}_{f_w^{-1} } \frac{1}{N}\sum_{k=1}^{K} \left[ \sum_{c\in \boldsymbol{o}_k}\left(f_w^{-1}(c) - \acute{c} \right)^2 \right] \text{,}
\label{eq:objective_objs}
\end{align}
where $N$ is the number of pixels in $I$. The sum of the differences of $\boldsymbol{o}_k$ can be reformulated by merging two summations in \eqref{eq:objective_objs}. Then, the distance is a sum of the squared difference between $f_w^{-1}(c)$ and $\acute{c}$. Converting \eqref{eq:notations_obj_delta_eps} into pixel-level representation and substituting $\acute{c}$ into its mean and deviations lead to
\begin{align}
   \operatorname*{arg\,min}_{f_w^{-1} } \frac{1}{N}\sum_{c\in I}\left(f_w^{-1}(c) - \tilde{c}_{cls} -\Delta_{c,cls} \right)^2
\label{eq:objective_pixels}
\end{align}
where $\tilde{c}_{cls}$ is an element of $\tilde{\boldsymbol{o}}_{k,cls}$. Since the objective function \eqref{eq:objective_objs} and \eqref{eq:objective_pixels} corresponds to the error criterion of the neural network, especially autoencoder, where the pixel differences can be calculated after forward-passing the input image, we now let $f_w^{-1}$ be an autoencoder network and train to infer $\acute{c}$ over the dataset S. Then we can expect that the trained autoencoder can restore the original image considering implicit semantic, $cls$. Thus, the objective function over the entire dataset S of $M$ equal-sized images is
\begin{align}
  \operatorname*{arg\,min}_{f_w^{-1} } \sum_{m=1}^{M}\left[
    \sum_{c\in I_m}\left(
      f_w^{-1}(c) - \tilde{c}_{cls} -\Delta_{c,cls}
    \right)^2
  \right] \text{.}
\label{eq:objective_over_s_1}
\end{align}
By definition, $\sum_{c\in S_{cls}}\Delta_{c,cls}=0$ where $S_{cls}$ is a subset of $S$ that belongs to a class label $cls$. With this definition, after some calculation, eq.~\eqref{eq:objective_over_s_1} becomes
\begin{align}
  \operatorname*{arg\,min}_{f_w^{-1} } \sum_{m=1}^{M} \left[
  \sum_{c\in I_m} \left( \left(f_w^{-1}(c) - \tilde{c}_{cls} \right)^2 + 2\epsilon_c \Delta_{c,cls} - \Delta_{c,cls}^2 \right) \right]  \text{.}
\label{eq:objective_over_s_2}
\end{align}
Since $\Delta^2$ is a constant for a given dataset $S$, the minimization becomes 
\begin{align}
  \operatorname*{arg\,min}_{f_w^{-1} } \sum_{m=1}^{M} \left[
  \sum_{c\in I_m} \left( \left(f_w^{-1}(c) - \tilde{c}_{cls} \right)^2 + 2\epsilon_c \Delta_{c,cls} \right) \right] \text{.}
\label{eq:objective_over_s_3}
\end{align}
Then $f_w^{-1}$ learns to restore image toward the mean, between the mean and original. Therefore, one can geometrically assume that $\epsilon_c$ at the optimum point is smaller than and proportional to $\Delta_{c,cls}$. Since deducing $\Delta_{cls}$ solely from a single image is an ill-posed problem, our method minimizes the influence of the inevitable error $\epsilon_c$ by collecting pixels during regression in chapter 4.1.

\section{Filter Style Estimation}

\subsection{Filter Parameterization}

Most of the image filtering and editing can be built with three operations – brightness, contrast, and color controls. Even though there are some local operations such as Vignetting, for simplicity, we have not considered those in this study. In general the three primary operations can be expressed as linear or polynomial functions for input image $x$ and output image $y$;
\begin{alignat}{2}
  &\text{Brightness} \qquad\qquad && y_1=x+c \label{eq:brightness} \\
  &\text{Contrast} && y_2=ax+b \label{eq:contrast} \\
  &\text{Color} && y_3=\sum_{i=1}^{\alpha}\left( e_ix_i^3+f_ix_i^2+g_ix_i \right) \label{eq:color}
\end{alignat}
where $\alpha$ is the number of color channels, three (i.e. RGB) in our case. After adding up $y_1$, $y_2$, and $y_3$ and expressing parameters as $\boldsymbol{\beta}$, the three operations becomes
\begin{align}
  y_{\gamma}= \beta_{\gamma, 0} + \sum_{i=1}^{3}\left( \beta_{\gamma,i1}x_i +\beta_{\gamma,i2}x_i^2 +\beta_{\gamma,i3}x_i^3 \right) \text{.}
\label{eq:global_editing_op}
\end{align}
Note that we repeatedly calculate $y_{\gamma}$ over the output color channel, i.e. $\gamma \in \{R, G, B\}$. Hereafter, we omit $\gamma$ for brevity.  Since \eqref{eq:global_editing_op} represents global editing operations, using this, we model the parametric function $f_{\boldsymbol{\beta}}^*$ of $f_{w}$ in the regarding $\boldsymbol{\beta}$. Thus, with the original and the reference image, we can approach filter parameter extraction, obtaining $\boldsymbol{\beta}$ in \eqref{eq:global_editing_op}, as a nonlinear regression problem operated at every pixel of an image. Hence the minimization target $E(\boldsymbol{\beta})$ is
\begin{align}
  E(\boldsymbol{\beta}) & = \sum_{n=1}^{N} \left( y_n - f_{\boldsymbol{\beta}}^*(\acute{x}_n) \right)^2
\label{eq:editing_op_minimization_true_1}
\end{align}
where $(\acute{x}, y)$ is a color pair in $(\acute{I}, I)$. After applying \eqref{eq:notations_obj_delta_eps}, \eqref{eq:editing_op_minimization_true_1} becomes \eqref{eq:editing_op_minimization_true_2}. In a normalized color domain, recalling the geometrical interpretation of \eqref{eq:objective_over_s_3}, $\epsilon$ becomes small, and thus the high order of $\epsilon$ becomes negligibly small. After some calculation and with the assumption that $\epsilon$ is proportional to $\Delta$, \eqref{eq:editing_op_minimization_true_2} becomes
\begin{align}
  E(\boldsymbol{\beta}) & = \sum_{n=1}^{N} \left( y_n - f_{\boldsymbol{\beta}}^*( f_w^{-1}(y_n) + \epsilon_n ) \right)^2
\label{eq:editing_op_minimization_true_2} \\
  & \approx \sum_{n=1}^{N} \left( y_n - f_{\boldsymbol{\beta}}^*( f_w^{-1}(y_n) ) \right)^2 \text{.}
\label{eq:editing_op_minimization}
\end{align}
Since high order terms of $\epsilon_n$ are ignored, \eqref{eq:editing_op_minimization} is a rough approximation on a single image. Due to the ill-posed nature of the problem, instead, we propose uncertainty-based regression in chapter 4.2 to alleviate the error of rough approximation. To this end, we set the problem to weighted least squares and solved it using quasi-Newton optimization. Note that eq.~\eqref{eq:editing_op_minimization} works well when $\sum_{c\in I}\Delta_{c,cls}$ is close to $0$.

\begin{figure}
\centering
\includegraphics[width=\linewidth]{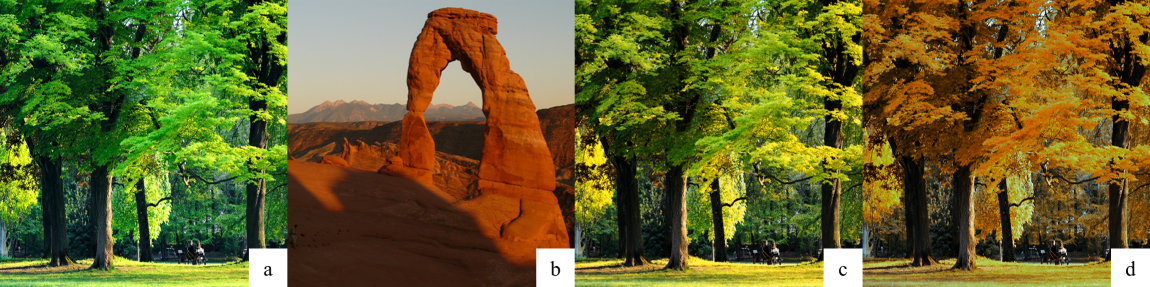}
\caption{{\bf The result comparison with and without correlation term.} (a) is an input image. (b) is a stylized reference image. (c) and (d) are FST results using \eqref{eq:global_editing_op} and \eqref{eq:global_editing_op_cc}, respectively.}
\label{fig:fig3}
\end{figure}

Additionally, our method can also provide results similar to color transfer, depending on filter parameters. If there are insufficient samples in the RGB color domain in a stylized image, it would not be easy to infer the coefficients to cover the absence of samples. In this case, if the Channel Correlation term (CC) is added, those colors can be transferred to other colors correlated with RG, RB, GB values as below.
\begin{align}
  y_{\gamma}= \beta_{\gamma, 0} + \sum_{i=1}^{3}\left( \beta_{\gamma,i1}x_i +\beta_{\gamma,i2}x_i^2 +\beta_{\gamma,i3}x_i^3 \right) + \beta_{\gamma,1}x_1x_2 + \beta_{\gamma,2}x_1x_3 + \beta_{\gamma,3}x_2x_3 
\label{eq:global_editing_op_cc}
\end{align}
Then the result becomes more like color transfer than FST. For example, in Fig.~\ref{fig:fig3}, green color is absent in stylized image. Therefore, green is transferred to another color in the result. More details will be presented in chapter 5.4.

\subsection{Uncertainty-based Adaptive Filter Regression}

After training $f_w^{-1}$, for any single filtered style image, we depend on a trained neural network to get the restored image and to regress the approximate function $f_{\boldsymbol{\beta}}^*$. In the process, there are high order terms of $\epsilon$ that causes the filter estimation error in the previous chapter. Due to the lack of evidence to directly minimize the error, we propose a roundabout method using the uncertainty of the inference and lower the weight where $\epsilon$ is expected high.

The error term is inherently non-negligible in our case since the function inferred $\Delta_{c,cls}$ solely from a single image. To be more specific, in the single inference of $I$, from each pixel $c$, the uncertainty of $f_w^{-1}(c)$ would be high when the implicit class variance $Var(\Delta_{c,cls})$ is high over the entire set $S$ (aleatoric uncertainty). That means if the variance of the deviation of $cls$ is high over $S$, the function $f_w^{-1}(c)$ is likely to give larger $\epsilon$. Moreover, $f_w^{-1}(c)$ would be more uncertain as $\Delta_{c,cls}$ is increased (epistemic uncertainty). In this case, the weight of unsure pixels should be lower when regressing the approximate function, $f_{\boldsymbol{\beta}}^*$.

The error term is independent over pixels. Therefore, we compute variance term same as combined uncertainty, a combination of epistemic uncertainty from Mean Standard Deviation (Mean STD) in the earlier study~\cite{gal2017deep} and aleatoric uncertainty in \cite{kendall2017uncertainties} in recovering the original image. Then for every pixel, the inverse of uncertainty, written as $\boldsymbol{\Omega}^{-1}$, is used as a weight of the least squares criterion. Then the general form of the solution to our regression problem is
\begin{align}
  \boldsymbol{\beta} = (\boldsymbol{X}^T\boldsymbol{\Omega}^{-1}\boldsymbol{X})^{-1}\boldsymbol{X}^T\boldsymbol{\Omega}^{-1}\boldsymbol{y}
\label{eq:beta_uncertainty}
\end{align}
In practice, our uncertainty-based regression can be achieved by multiplying $\boldsymbol{\Omega}^{-1/2}$ to both of the $\boldsymbol{X}$ and $\boldsymbol{y}$ followed by quasi-Newton optimization. Note that $\boldsymbol{X}$ and $\boldsymbol{y}$ are the design matrix that consists of stacked polynomial vectors of the restored original colors and the vector of the filtered colors, respectively.

\subsection{Regularization}

\begin{figure}
\centering
\includegraphics[width=5cm]{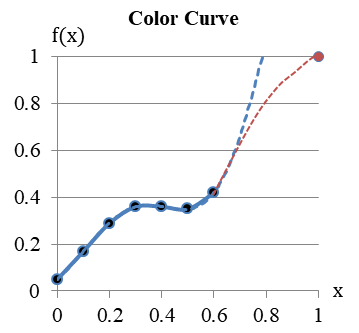}
\caption{{\bf An example of a nonlinear function that has unfavorable extrema matching in a color transfer curve.} Since there is no color sample near $x = 1$, the function can diverge as shown by the blue dotted line. Instead, we can add a regularization term to guide nonlinear function into a red line.}
\label{fig:fig4}
\end{figure}

With the methods described in previous chapters, we are now able to estimate the parameters of filter style from single image input. However, there are two problems with this unrefined algorithm. Firstly, when we do not have enough plots around each extremum of color space, the regression function is left to vary dramatically outside of plots. Like the blue line shown in Fig.~\ref{fig:fig4}, a polynomial function curves very fast without the basis of plots, leading the extrema transformed to unfavorable values. So, for the new user input image $X$, stylized image $Y$ often show clipping or extreme colors around the extremes. The second problem is that the output can sometimes be visually unnatural as the regression function severely deviates from linear, leaving the regression process vulnerable to specific colors, which are exceptionally scarce but saturated in the pairs of the stylized and inferred original. 

To relieve the above symptoms, we design and add regularization term in the regression function. To deal with the first phenomenon, we regularize the function to be close to $(0, 0)$, $(1, 1)$, respectively, when the color range is normalized. Hence, an L2 penalty is added and penalizes the function when it starts to diverge from 1 and 0. For each output channel $\gamma$, 
\begin{align}
  R_{1,\gamma}= \beta_{\gamma,0}^2 + \left\{ \left[ \beta_{\gamma,0} + \sum_{i=1}^{3}\left( \beta_{\gamma,i1} + \beta_{\gamma,i2} + \beta_{\gamma,i3} \right) \right] -1 \right\}^2
\label{eq:regularization_1}
\end{align}
After adding \eqref{eq:regularization_1}, the nonlinear function looks like a red line instead of a blue line in Fig.~\ref{fig:fig4}.

\begin{figure}
\centering
\includegraphics[width=\linewidth]{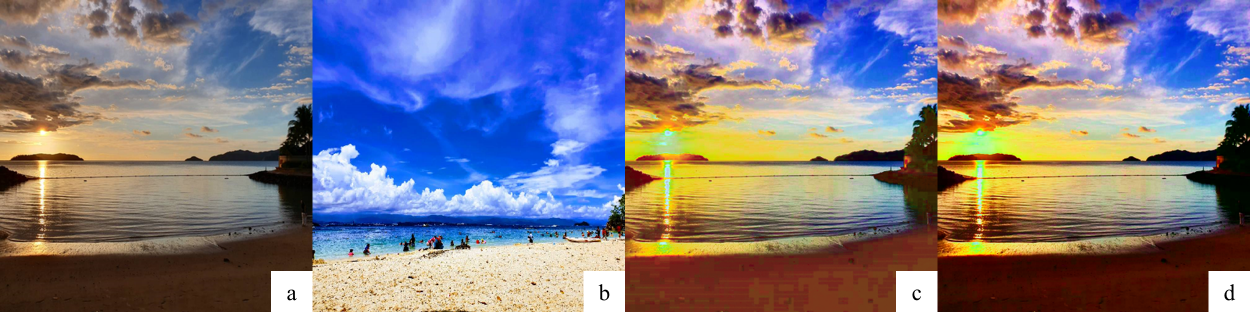}
\caption{{\bf An example of the regularization effect.} Regularization term yields a stable result where there is less or no clipping. (a) is an input image, (b) is a filtered style image, and (c) and (d) are results of FST without and with regularization terms, respectively.}
\label{fig:fig5}
\end{figure}

For the second symptom, we add an L2 penalty on the coefficients of high order terms. Empirically we found out that imposing the L2 penalty only on different sources of colors is visually good, rather than imposing L2 on all color sources.
\begin{align}
  R_{2,\gamma}=  \sum_{i\neq \gamma}\left( \beta_{\gamma,i2}^2 + \beta_{\gamma,i3}^2 \right)
\label{eq:regularization_2}
\end{align}
After adding two regularization terms \eqref{eq:regularization_1} and \eqref{eq:regularization_2}, the error of regression function becomes
\begin{gather}
  \acute{E}(\boldsymbol{\beta}_\gamma) = E(\boldsymbol{\beta}_\gamma) + \lambda R(\boldsymbol{\beta}_\gamma) \text{, where} \\
  R(\boldsymbol{\beta}_\gamma)= \beta_{\gamma,0}^2 + \left\{ \left[ \beta_{\gamma,0} + \sum_{i=1}^{3}\left( \beta_{\gamma,i1} + \beta_{\gamma,i2} + \beta_{\gamma,i3} \right) \right] -1 \right\}^2 + \sum_{i\neq \gamma}\left( \beta_{\gamma,i2}^2 + \beta_{\gamma,i3}^2 \right)
\end{gather}
In Fig.~\ref{fig:fig5}, we show the difference in result images by introducing our regularization term. In this figure, the result when regularization weight is $0$ shows clipping around the ground region, while the full use of the regularization does not exhibit this behavior. Note that $\lambda$ can be obtained by grid-search.

\section{Experiment}

\subsection{Dataset}

To generalize the defilterization network $f_w^{-1}$ for various scenes, we require a large-scale dataset with lots of classes. One of the most popular image datasets, MSCOCO~\cite{lin2014microsoft}, is widely used and contains more than 110K images with 80 object categories. However, this dataset does not contain filtered images. Therefore, to generate filtered images, we applied various types of real and arbitrary synthetic photo filters to the dataset. Initially, a filtered dataset was generated by posing 26 real Instagram filters using publicly available source code in CSSgram~\cite{Unacssgram}. In addition, synthetic filtered images were added by posing random color, contrast, and brightness six times. Based on this dataset, we trained the defiliterization network and tested our proposed method. 

To check the dataset dependecy, we also prepared 99 private photos and 17 unseen real filters from one of the camera application in Android Play Store.
As shown in the experiment in the next chapter, the proposed method, FST, can successfully transfer filter effects from a single image to the new input, even the filters unseen in the training phase.

\subsection{Evaluation}

Firstly we prepared the architecture of image-to-image translation, introduced in \cite{johnson2016perceptual}, as a defilterization function $f_w^{-1}$. Then, we trained $f_w^{-1}$ using the combined dataset and fully synthetic dataset (self-supervised learning) until the test MSE is saturated. Note that the filter transferred output would show better results with a better choice of defilterization network and more synthetic dataset generation, but we leave it as further work.

\begin{figure}
\centering
\includegraphics[width=\linewidth]{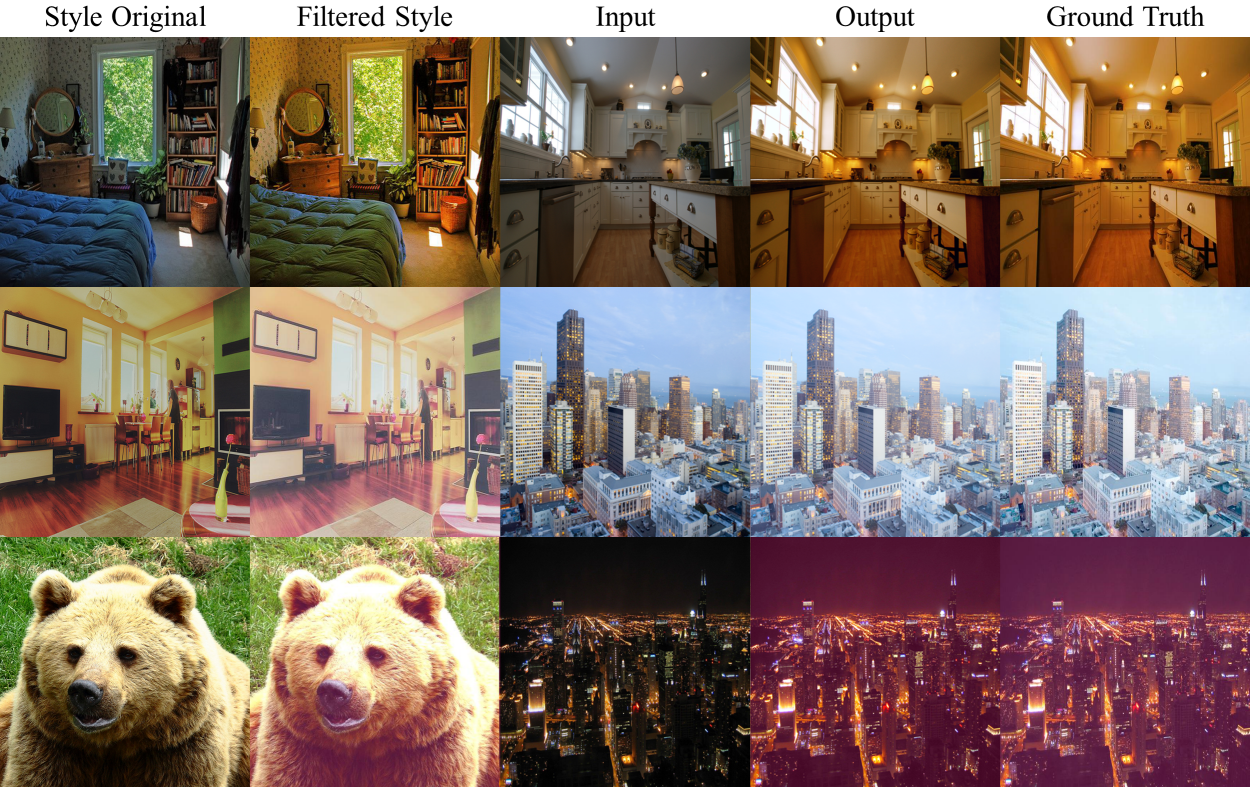}
\caption{{\bf Qualitative results of FST.} We also included the originals of filtered style images, which were not given during inference.}
\label{fig:fig6}
\end{figure}

To validate our proposed method, we randomly chose 100 images from the MSCOCO validation set, and selected 18 real filters out of 26 filters to generate 1800 filtered images. We excluded eight filters that have a noticeable vignetting effect, which is outside of the scope of the current study. Then, FST was performed on the remaining validation images, and the result was compared with the ground truth images, which were directly generated by applying filters. Quantitative and qualitative results are given in Table~\ref{table:quantitative_results} and Fig.~\ref{fig:fig6}, respectively. To further test our method on unseen dataset, we also evaluated the proposed method on private photos with 43 filters (17 unseen filters and 26 Instagram filters). The result is given in Table~\ref{table:quantitative_results2}.

\setlength{\tabcolsep}{4pt}
\begin{table}
\begin{center}
\caption{{\bf Quantitative evaluation.} The MSCOCO validation set with 18 Instagram filters is used for evaluation. Our method supports a few variations. Our method can have (R): regularization, (AU): Aleatoric uncertainty, (U): combined uncertainty, (CC): color correlation. Note that for WCT2, we gave option that uses features from decoder and skip-connection since it performs the best. Note that lower $\Delta E_{00}^*$ (a.k.a., Delta-E 2000) is better.}
\label{table:quantitative_results}
\begin{tabular}{lll}
\hline\noalign{\smallskip}
Methods & PSNR & $\Delta E_{00}^*$\\
\noalign{\smallskip}
\hline
\noalign{\smallskip}
{\bf Ours } & {\bf 25.226 } & {\bf 6.660 } \\
{\bf Ours (w. U, CC) } & {\bf 24.931 } & {\bf 6.725 } \\
{\bf Ours (w. AU) } & {\bf 25.438 } & {\bf 6.427 } \\
{\bf Ours (w. U) } & {\bf 25.495 } & {\bf 6.394 } \\
{\bf Ours (w. R, U) } & {\bf 26.093 } & {\bf 6.148 }\\
WCT2 \cite{yoo2019photorealistic} & 16.473 & 17.516 \\
Color Transfer \cite{reinhard2001color} & 7.325 & 34.914 \\
\hline
\end{tabular}
\end{center}
\end{table}
\setlength{\tabcolsep}{1.4pt}

\setlength{\tabcolsep}{4pt}
\begin{table}
\begin{center}
\caption{{\bf Quantitative evaluation on private photos with 43 filters.}}
\label{table:quantitative_results2}
\begin{tabular}{lllllllll}
\hline\noalign{\smallskip}
Methods & Ours (w. R, AU) & Ours (w. R, U) & WCT2~\cite{yoo2019photorealistic} & Color Transfer~\cite{reinhard2001color}\\
\noalign{\smallskip}
\hline
\noalign{\smallskip}
PSNR & {\bf 25.814 } & {\bf 25.850 } & 17.234 & 6.985 \\
$\Delta E_{00}^*$ & {\bf 5.881 } & {\bf 5.854 } & 15.723 & 35.123 \\
\hline
\end{tabular}
\end{center}
\end{table}
\setlength{\tabcolsep}{1.4pt}

\subsection{Comparison with Style Transfer}

Our problem definition is inherently different from conventional style transfer researches. Style transfer seeks the transfer of texture, color, and even abstract concepts, while our method targets the transfer of photo editing or filter effects applied to an image. Although photorealistic stylization approaches~\cite{li2018closed, yoo2019photorealistic} show attractive results in terms of structure preservation, they tend to directly transfer color distribution from the style to content images, not transferring filter information of the style image. Nonetheless, we compared our proposed FST with photorealistic style transfer, since style transfer is the most similar task.

\begin{figure}
\centering
\includegraphics[width=\linewidth]{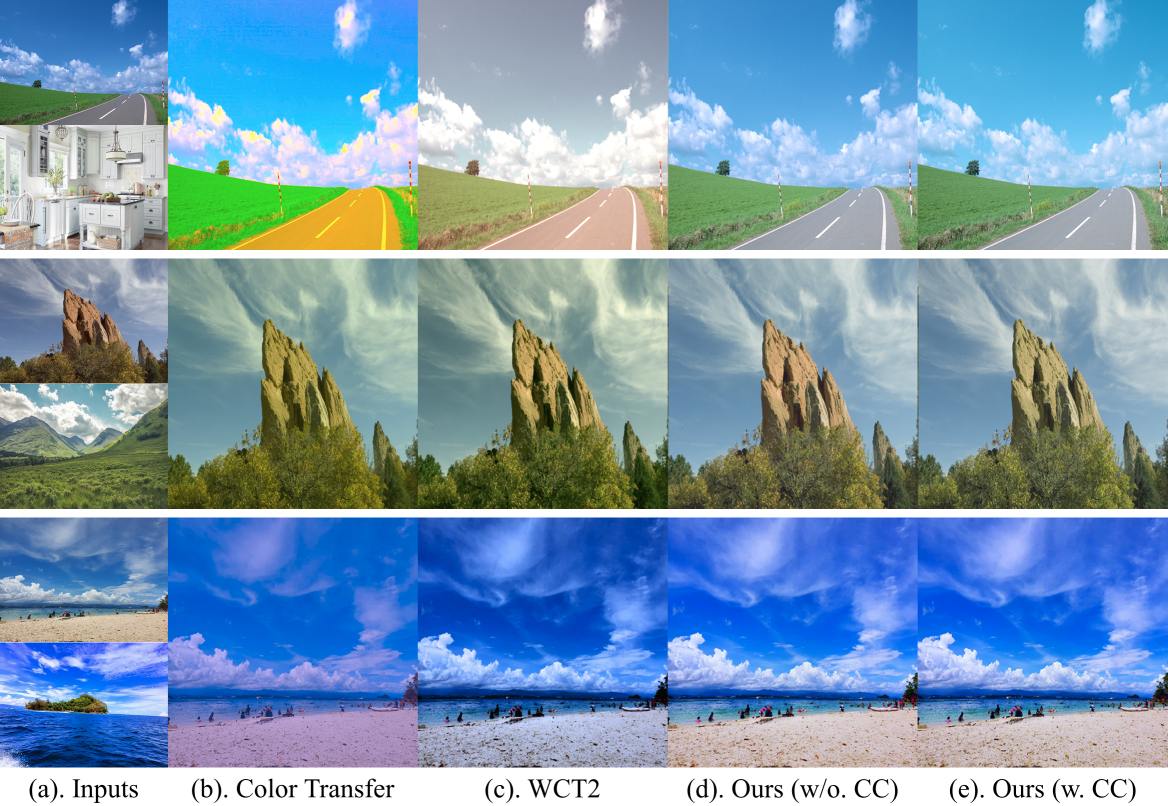}
\caption{{\bf The results of FST with channel correlation.} Given (a) an input pair (top: content, bottom: style), the results of (b) color transfer~\cite{reinhard2001color}, (c) WCT2~\cite{yoo2019photorealistic}, (d) and (e) ours (FST) are shown. With CC (color correlation) term, the proposed method can also transfer colors of contents, whitish color in the first row, green forests in the second row, and bluish color in the third row, for example.}
\label{fig:fig7}
\end{figure}

\begin{figure}
\centering
\includegraphics[width=\linewidth]{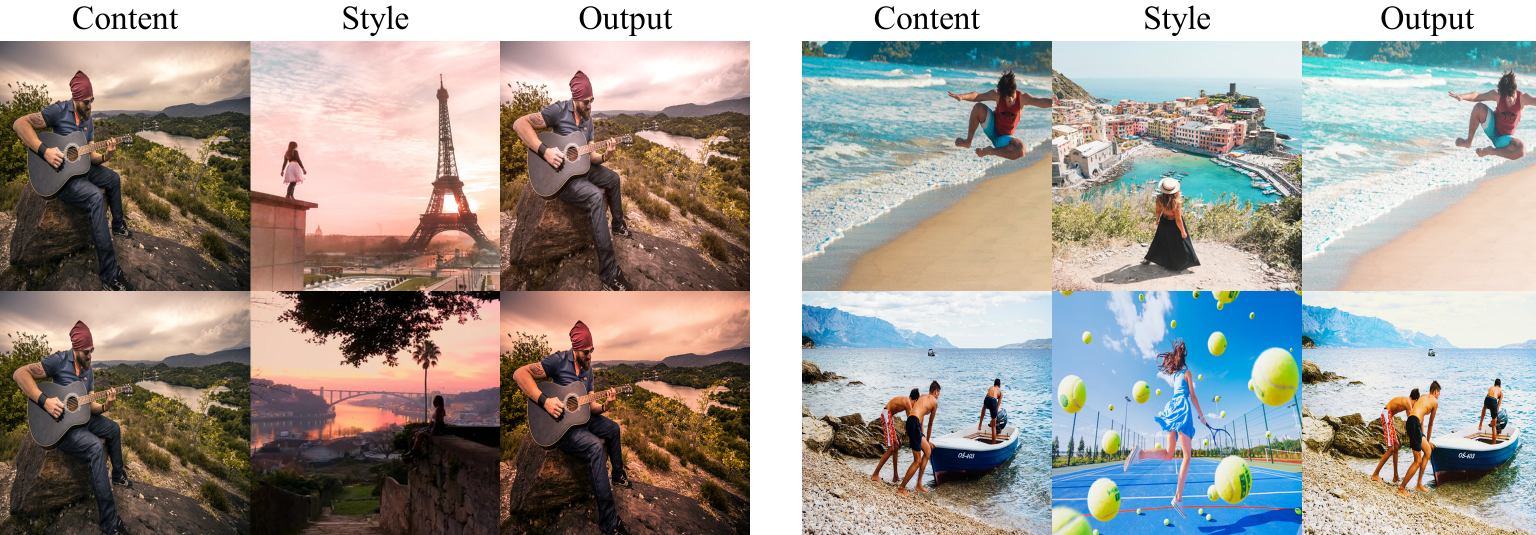}
\caption{{\bf More results from the proposed FST using stylish photos.} }
\label{fig:fig8}
\end{figure}

To validate the effectiveness of the proposed method, we compare it with two types of photorealistic style transfer methods based on conventional linear color distribution transfer~\cite{reinhard2001color} and structure-preserved style transfer based on a high-frequency component skip, which is the state-of-the-art in photorealistic style transfer~\cite{yoo2019photorealistic}. As shown in Fig.~\ref{fig:fig7}, both approaches tend to directly reflect the color distribution of the style image to the content image while FST transfers filter style only. In case of color transfer, if style image consists of achromatic colors mostly, it often generates visually unpleasing and questionable results. Furthermore, in terms of computational complexity, while conventional style transfer techniques take several hundred milliseconds, our solution can transfer filter style to the content image in less than a few milliseconds. Moreover, FST can transfer stylish effects as well as unseen filters. Detailed results are shown in Fig.~\ref{fig:fig8}.

\subsection{Application}

{\bf A trade-off between Filter and Color Style.} \space\space\space\space In chapter 4.1, the approximate filter-applying function is modeled as a polynomial form \eqref{eq:global_editing_op} to transfer filter styles. However, in addition to filter style, more dramatic effects may be required upon requests. In this case, by adding a correlation term to filter-applying function as shown in \eqref{eq:global_editing_op_cc}, FST can transfer some colors that are not present in the reference style image to neighbor colors at the expense of quantitative accuracy. As shown in Fig~\ref{fig:fig3}, results look more like color style transfer than the original FST. 

{\bf Real-time Filter Transfer on Mobile Device.} \space\space\space\space Along with the satisfactory results of our method, it is designed to run real-time on a mobile environment, where there is a severe restriction on computational power. FST takes most of its time on filter parameter extraction, and it costs as much as the inference time of autoencoder, plus nonlinear regression. However, once the parameters are obtained, transferring the filter style onto the new input can be done almost instantly. The processing times are firstly measured on a PC, and compared with WCT2 and Color Transfer. Results are given in Table~\ref{table:time_results}. Furthermore, in a mobile environment, our approach performs with 900ms on average on Qualcomm Snapdragon 855 to process an FHD image. The processing time on a smartphone is given in Table~\ref{table:time_result_mobile}. Note that once the filter is extracted, RGB Look Up Table (LUT) can be precomputed and stored on the device to shorten the processing time. Then, in the run-time, pixels are matched to new values using LUT to generate a stylized output image, which requires less than 2ms to transfer FHD images.
\setlength{\tabcolsep}{4pt}
\begin{table}
\begin{center}
\caption{{\bf Run-time comparison on a PC.} Steps are divided into two; filter extraction and application. Tests were done using the machine with Python Numpy, Nvidia GTX 1080ti, and Intel i7-8700 CPU. We used 256x256 image for filter extraction and 1920x1080 (FHD) for filter application. Note that excluding Epistemic Uncertainty (EU) in the proposed method can shorten the time required for filter extraction. We used 10 MC dropouts for EU.}
\label{table:time_results}
\begin{tabular}{lll}
\hline\noalign{\smallskip}
Methods & Filter extraction & Filter application\\
\noalign{\smallskip}
\hline
\noalign{\smallskip}
{\bf Ours (w/o. EU) } & {\bf 28ms } & {\bf 86ms } \\
{\bf Ours (full ver.) } & {\bf 132ms } & {\bf 86ms }\\
WCT2 \cite{yoo2019photorealistic} & 16ms & 943ms\\
Color Transfer \cite{reinhard2001color} & \multicolumn{2}{c}{49ms (single stage)}\\
\hline
\end{tabular}
\end{center}
\end{table}
\setlength{\tabcolsep}{1.4pt}
\setlength{\tabcolsep}{4pt}
\begin{table}
\begin{center}
\caption{{\bf Run-time on a smartphone.} We also measured our method in a mobile environment, Samsung Galaxy S10. We used 256x256 image for filter extraction and 1920x1080 (FHD) for filter application. The filter application is much faster than the PC version since the mobile version uses GPU parallel processing, while the PC version partially uses Numpy CPU. Note that we excluded EU from the mobile version.}
\label{table:time_result_mobile}
\begin{tabular}{lll}
\hline\noalign{\smallskip}
Methods & Filter extraction & Filter application\\
\noalign{\smallskip}
\hline
\noalign{\smallskip}
{\bf Ours, mobile version } & {\bf 900ms } & {\bf 2ms } \\
\hline
\end{tabular}
\end{center}
\end{table}
\setlength{\tabcolsep}{1.4pt}

\section{Conclusion and Future Work}

To the best of our knowledge, this is the first study on the real-time filter transfer between two real photos. Our approach resolves a new task called FST (Filter Style Transfer), transferring custom filter operation between images, which is different from previous works of style transfer. Although style transfer methods yield reasonable outputs in some cases, it does not consistently generate pleasant outputs in every case and may require an additional effort on tuning the result. Moreover, arbitrary photorealistic style transfer still provides degraded or ruined texture, which is not desirable in photo filter extraction.

In a mobile environment, once filter parameters acquired, FST consistently runs within 2ms to transfer FHD previews in camera applications, which shows exceptional real-time performance. Moreover, our solution can also perform similar to color transfer depending on the regression function, but it still creates more natural output than any other existing color transfer method.

In the proposed method, a defilterization network can be replaced by any other network structures. However, note that the performance of the network directly relates to the performance of filter transfer. In the future, we therefore plan to work on improving the performance of the defilterization network to extract the originals from filtered images more accurately. For implicit learning of semantical objects, it may require to train class labels explicitly as well as decoder part. If the defilterization network can perfectly extract the original image, the final result of the filter transfer would be more reliable.

%
%
\bibliographystyle{splncs04}
\bibliography{FST}


\end{document}